\documentclass[final]{cvpr}

\usepackage{times}
\usepackage{epsfig}
\usepackage{graphicx}
\usepackage{amsmath}
\usepackage{amssymb}
\usepackage{booktabs}
\usepackage{caption}
\usepackage{subcaption}
\usepackage{makecell}
\usepackage{multirow}

\usepackage[pagebackref=true,breaklinks=true,colorlinks,bookmarks=false]{hyperref}

\def\cvprPaperID{6198} 
\def\confYear{CVPR 2021}

\begin{document}

\title{Rotation-Invariant Autoencoders for Signals on Spheres}

\author{Suhas Lohit\\
Mitsubishi Electric Research Laboratories\\
Cambridge, MA\\
{\tt\small slohit@merl.com}
\and
Shubhendu Trivedi\\
Massachusetts Institute of Technology\\
Cambridge, MA\\
{\tt\small shubhendu@csail.mit.edu}
}

\maketitle

\begin{abstract}
   Omnidirectional images and spherical representations of $3D$ shapes cannot be processed with conventional 2D convolutional neural networks (CNNs) as the unwrapping leads to large distortion. Using fast implementations of spherical and $SO(3)$ convolutions, researchers have recently developed deep learning methods better suited for classifying spherical images. These newly proposed convolutional layers naturally extend the notion of convolution to functions on the unit sphere $S^2$ and the group of rotations $SO(3)$ and these layers are equivariant to 3D rotations. In this paper, we consider the problem of unsupervised learning of rotation-invariant representations for spherical images. In particular, we carefully design an autoencoder architecture consisting of $S^2$ and $SO(3)$ convolutional layers. As 3D rotations are often a nuisance factor, the latent space is constrained to be exactly invariant to these input transformations. As the rotation information is discarded in the latent space, we craft a novel rotation-invariant loss function for training the network. Extensive experiments on multiple datasets demonstrate the usefulness of the learned representations on clustering, retrieval and classification applications.
\end{abstract}

\section{Introduction}
\label{sec:intro}
Advances in computational imaging and robotics has resulted in an increase in interest in processing spherical/omnidirectional images obtained from $360^0$ cameras~\cite{yagi1999omnidirectional,gluckman1998ego}. As such cameras have been becoming cheaper, they are now part of products such as robots~\cite{paya2017state}, surveillance cameras~\cite{morita2003networked} and ``action cameras" such as GoPros. However, conventional deep learning techniques for regular images i.e. 2D convolutional neural networks (CNNs) which are the best performing machine learning methods of today, are not suitable for processing spherical images. This is because, in order to employ the regular 2D CNN, the signal on the sphere must be first unwrapped into a 2D array, which creates a large amount of non-uniform distortion througout the image (the topology of $S^2$ is different from that of $\mathbb{R}^2$). The same is true for the case of spherical representations of 3D shapes, where the 3D shapes are converted to functions on spheres using ray-casting. This incompatibility between domains necessitates novel designs of neural networks which are tailored to spherical images. 

To this end, Cohen et al. \cite{cohen2018spherical} and Esteves et al. \cite{esteves2018learning} independently designed convolutional neural networks that can operate directly on signals on spheres. Both these methods involve defining a correlation function between signal on a sphere and a trainable kernel, which is also a signal on the same sphere. This can be achieved computationally efficiently using spherical harmonics. An important feature of this correlation operation is that it is \textit{equivariant} to 3D rotations of the sphere. This is analogous to the usual 2D CNNs for regular images which are equivariant to 2D translations. These spherical convolutional layers can then be stacked along with non-linearities to form deep neural networks for spherical signals. These networks have so far mainly been employed for classification problems where a spherical image is mapped to one of many predefined classes.

All these networks are trained in a supervised fashion, and require a lot of labeled data. In this paper, we propose an unsupervised framework for spherical images. In particular, we design a novel autoencoder architecture such that the latent features remain exactly invariant to any rotation of the input signal. This is because, in many applications, \textit{the inputs do not have a canonical rotation, and the particular orientation that the input signal is measured in is a nuisance factor and should ideally be factored out}. However, as the latent space is rotation-invariant, the usual loss functions cannot be employed for training the proposed autoencoder. Instead, we design a novel loss function, based on the spherical cross-correlation between the estimated output and the desired output, which is also invariant to the relative orientation of the estimated output with respect to the desired output. We now summarize our contributions below. 

\begin{enumerate}
    \item We propose a novel autoencoder architecture for spherical images, where the latent features are guaranteed to be invariant to 3D rotations applied at the input.
    \item We carefully design a novel rotation-invariant loss function, based on the maximum value of the spherical cross-correlation between the estimated and desired outputs which is used to train the rotation-invariant spherical autoencoder.
    \item We show that the latent features learned using the proposed unsupervised framework are superior to a vanilla autoencoder which does not use the proposed rotation-invariant loss function, for multiple tasks including clustering, retrieval and classification.
\end{enumerate}

\section{Related Work}
\label{sec:related_work}
In this section, we briefly review literature related to invariant representations using deep learning and equivariant neural networks. Before that, we define equivariance and invariance with respect to a set of transformations $\mathcal{T}$. Consider, for simplicity, functions $f, g : L^2(\mathbb{X}) \to L^2(\mathbb{X})$. For all $x \in \mathbb{X}$, where $\mathbb{X}$ is some space on which the functions are defined. $f(.)$ is invariant with respect to $\mathcal{T}$ if $\forall T \in \mathcal{T}, f \circ T(x) = f(x)$. Similarly, $g(.)$ is equivariant with respect to $\mathcal{T}$ if $\forall T \in \mathcal{T}, g \circ T(x)  = T \circ g(x)$. 

\subsection{Invariant representations using deep learning}
While deep learning has produced some very good results for various applications in computer vision, they come at the expense of requiring large amounts of training data and computational resources \cite{krizhevsky2012imagenet} and a disadvantage that these deep models are often very sensitive to small input variations \cite{dodge2016understanding}. At the same time, these overparameterized models are prone to learning mappings that are spurious correlations \cite{arjovsky2019invariant}, also known as shortcuts \cite{geirhos2020shortcut}. Goodfellow et al. \cite{goodfellow2009measuring} showed that while regular CNNs do provide some translational invariance, invariance to other transformations cannot be guaranteed. To remedy this drawback, various approaches have been studied in the recent past. These include \textit{hybrid} model- and data-driven approaches such as spatial \cite{jaderberg2015spatial} and temporal transformers \cite{lohit2019temporal} which use specially designed modules to model nuisance factors, capsule networks for autoencoding by Hinton et al.~\cite{hinton2011transforming} and for classification by Sabour et al. \cite{sabour2017dynamic} which explicitly model pose variations, and other works which modify the convolutional layers for added scale \cite{xu2014scale} and affine invariances \cite{xu2020towards}. In the realm of unsupervised representation learning, which is the focus of this paper, several disentangling methods have been proposed. In these methods, the elements of latent vectors are encouraged to represent disentangled factors of variation. That is, if an element of the latent vector, which encodes a particular factor of variation, is removed, the rest of the vector remains invariant to that factor. Examples in computer vision include the works by Kulkarni et al. \cite{kulkarni2015deep} which disentangles factors of variation for face images, Shukla et al. \cite{shukla2019product} who propose using orthogonal latent spaces to encourage disentanglement,  Shu et al.~\cite{shu2018deforming} and Koneripalli et al. \cite{koneripalli2020rate} who use spatial and temporal transformer modules in autoencoders in order to encourage affine-invariant representations for images and rate-invariant representations for human motion sequences. Permutation-invariant unsupervised representations for point-clouds has also been effective~\cite{achlioptas2018learning,yang2018foldingnet}.
 
 \vspace{-0.1in}
\subsection{Equivariant neural networks}
In addition to the above methods, a new promising class of architectures has emerged which, unlike previous methods, can \textit{guarantee exact invariance} to certain transformations by design. The general architecture is as follows. There is a stack of \textit{equivariant} layers which operate on the input, followed by a pooling/aggregation layer which promotes the representations to be fully invariant. In the case of classification architectures, the invariant features are then passed through the layers of a classifier head.

As mentioned in the introduction, the CNNs used for regular images are equivariant to 2D translations. Cohen and Welling~\cite{cohen2016group} described Group-CNNs which are equivariant to the action of discrete rotations on images. This idea has been applied for autoencoders by Kuzminykh et al.~\cite{kuzminykh2018extracting}. These ideas were then extended to the case of continuous rotations by Weiler et al.~\cite{weiler2018learning} and to both rotations and translations by Worrall et al.~\cite{worrall2017harmonic}. The idea of CNNs have been extended for the case of spherical images by several authors. Kondor and Trivedi~\cite{kondor2018generalization} showed theoretically that a convolutional structure is actually necessary, and not just sufficient, in order to guarantee equivariance to actions of compact groups on signals. Cohen et al.~\cite{cohen2018spherical} and Esteves et al.~\cite{esteves2018learning} concurrently developed spherical CNNs based on spherical correlation layers, which are computed efficiently using the spherical and SO(3) Fourier transform (SOFT) \cite{kostelec}. Kondor et al. proposed Clebsch-Gordan Nets~\cite{kondor2018clebsch} which operated fully in the Fourier domain and is computationally more efficient. More recently, Esteves et al.~\cite{esteves2020spin} designed spin-weighted CNNs which are computationally more efficient compared to \cite{cohen2018spherical} and at the same time, do not sacrifice representation power unlike the isotropic filters used in \cite{esteves2018learning}. In this paper, we employ the layers proposed by Cohen et al.~\cite{cohen2018spherical}, however the ideas presented in this paper are applicable to these other architectures as well. Similar ideas have be used to design network layers equivariant to scaling and blurring operators on images ~\cite{worrall2019deep}, rotation equivariance for point clouds~\cite{zhang20193d}, learning on sets ~\cite{ravanbakhsh2016deep,zaheer2017deep,maron2020learning} and graphs ~\cite{kondor2018covariant,maron2018invariant,thiede2020general}, gauge equivariant transforms for spheres~\cite{cohen2019gauge} and meshes~\cite{de2020gauge}. Existing literature on equivariant networks has mainly focused on supervised learning and classification problems. \textit{In this paper, we propose an spherical CNN-based autoencoder for spherical images with a rotation-invariant latent space.} We now describe the background necessary to develop this framework.

\section{Background: correlations on $S^2$ and $SO(3)$}
\label{sec:background}
In this section, we provide a brief overview of the mathematical ideas required to define and compute correlations of signals defined on the unit sphere $S^2$ and the group of 3D rotations $SO(3)$. Please refer to the excellent article by Esteves \cite{esteves2020theoretical} for a longer treatment of these ideas. Briefly, by the unit sphere $S^2$, we mean the set of points $x \in \mathbb{R}^3$ such that $||x||_2^2 = 1$. Clearly, this is a 2-dimensional space and can be equipped with a Riemannian manifold structure with the usual inner product as the Riemannian metric. As such, the points on the sphere can be specified using a two-dimensional spherical co-ordinate system say $\alpha \in [0, 2\pi], \beta \in [0, \pi]$. The set of 3D rotations, denoted by $SO(3)$, is a three-dimensional space which can be specified using the popular convention of ZYZ Euler angles given by $\alpha \in [0, 2\pi], \beta \in [0, \pi], \gamma \in [0, 2\pi]$, which quantify the rotation around the Z-, X- and Z- axes respectively performed in order. Note that several such conventions exist and we can easily transform the coordinates from one system to another. Signals/images on $S^2$ and $SO(3)$ can be described as functions $f: S^2 \to \mathbb{R}$ and $g: SO(3) \to \mathbb{R}$ respectively. Note that filters in the layers as well as the output feature maps are such functions. 

We can now define a correlation -- convolution, in deep learning parlance -- between a spherical image $f$ on the sphere with a filter $h$, which is also a function on the sphere as follows:

\begin{equation}
    \label{eq:corrs2}
    h \star f (R) = \int_{S^2} h(R^{-1}x)f(x)dx,
\end{equation}

where $\star$ refers to $S^2$ correlation, $R \in SO(3)$ is a rotation matrix, $dx$ is a rotation invariant measure of integration given by $d\alpha \sin(\beta)d\beta/4\pi$. Note that (a) by this definition of correlation, the output of the correlation operator for two signals on $S^2$ is a signal on $SO(3)$. (b) this definition is not unique. Equation \eqref{eq:corrs2} is followed in this paper following Cohen et al.~\cite{cohen2018spherical}. Esteves et al.~\cite{esteves2018learning} define a different correlation operator where the output is still a function on $S^2$. They show that their definition restricts the filters $h$ in \eqref{eq:corrs2}, which are eventually learned from data, to be isotropic/zonal filters. This can increase computational efficiency, but at the cost of reduced expressivity. 

Similarly, for a function $f$ and a filter $h$ on $SO(3)$, the $SO(3)$ correlation is defined as

\begin{equation}
    \label{eq:corrso3}
    h \star f (R) = \int_{SO(3)} h(R^{-1}X)f(X)dX, 
\end{equation}

where $\star$ now denotes correlation on $SO(3)$, $dX$ is the measure of integration given in ZYZ Euler angles by $d\alpha \sin(\beta)d\beta d\gamma/8\pi^2$. Note that these definitions of correlations are equivariant to the rotation group $SO(3)$. That is, if the function $f$ undergoes a rotation $R$, the correlation output undergoes the same rotation. Please refer Cohen et al.~\cite{cohen2018spherical} and Kondor et al.~\cite{kondor2018clebsch} for more details. 

Now, we discuss how to compute them efficiently using spherical and $SO(3)$ Fourier transforms. We compute the Fourier coefficients for a signal on $S^2$ by projecting it onto the spherical harmonics, the elements of which are denoted by $Y_m^l(x)$, with two indices: the degree $l \geq 0$ and order $m$ such that $-l \leq m \leq l$. The Fourier coefficients of a bandlimited signal $f$ with a bandwidth of $b$ are thus given by 

\begin{equation}
    \label{eq:fts2}
    \hat{f}_m^l = \int_{S^2} f(x) \Bar{Y}_m^l(x)dx.
\end{equation}

It also follows that the inverse Fourier transform, i.e., the synthesis function is given by 

\begin{equation}
    \label{eq:ifts2}
    f(x) = \sum_{l=0}^b \sum_{m=-l}^l \hat{f}_m^l Y_m^l(x).
\end{equation}

Similar to the above, in order to compute the $SO(3)$ Fourier Transform of a signal defined on the group of rotations $SO(3)$, we project it onto the \textit{irreducible representations} of the $SO(3)$ group, which are called Wigner-D matrices, the elements of which are denoted by $D_{m,n}^l$. $D_{m,n}^l$ has a dimension $2l+1 \times 2l+1$ and $m, n$ index the rows and columns of the matrix. The Fourier coefficients of a signal $f$ defined on $SO(3)$ with a bandwidth of $b$ is given by 

\begin{equation}
    \label{eq:ftso3}
    \hat{f}_{m,n}^l = \int_{SO(3)} f(x) \Bar{D}_{m,n}^l(x)dx.
\end{equation}

The inverse $SO(3)$ Fourier transform is given by 

\begin{equation}
    \label{eq:ifts2}
    f(R) = \sum_{l=0}^b (2l+1) \sum_{m=-l}^l \sum_{n=-l}^l \hat{f}_{m,n}^l D_{m,n}^l(R).
\end{equation}

Now that the required Fourier and inverse Fourier transforms have been defined, we can compute the correlation on $S^2$ between $f$ and $h$ using the correlation theorem:

\begin{equation}
    \label{eq:corrfts2}
    \widehat{h \star f}^l = \hat{f}^l \hat{h}^{l \dagger}.
\end{equation}

That is, we first compute the Fourier coefficients $\hat{f}^l$ and $\hat{h}^l$ for $l=0,\dots,b$, which are vectors. Then we compute the outer product which gives us the Fourier transform of the  correlation operator. Finally, we compute the inverse $SO(3)$ Fourier transform which gives us the desired output of the correlation, which is a function on $SO(3)$.

In a similar fashion, we can compute the $SO(3)$ correlation between a function $f$ and a filter $h$, both defined on $SO(3)$ using a very similar formula as in Equation \eqref{eq:corrfts2}, where we use the forward $SO(3)$ Fourier transform for computing $\hat{f}$ and $\hat{h}$, which are now block matrices. They are multiplied together and then we compute the inverse $SO(3)$ transform to yield the correlation output, also on $SO(3)$. In this paper, we use the excellent Pytorch/CUDA implementations of the above functions provided by Cohen et al.~\cite{cohen2018spherical} for building our framework.

\section{Rotation-invariant autoencoder for spherical images}
\begin{figure}
    \centering
    \includegraphics[width=\linewidth]{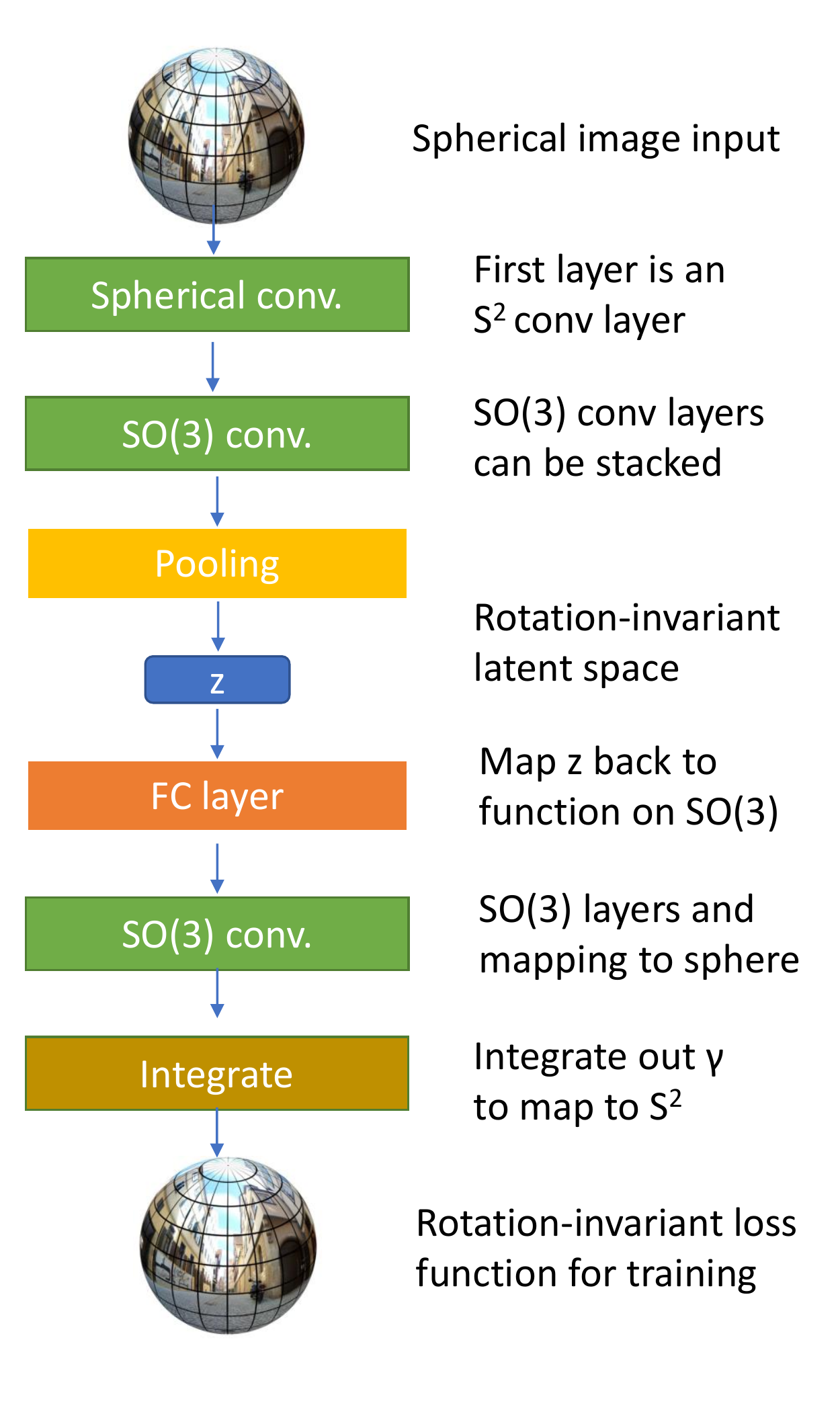}
    \caption{The figure provides an overview of the proposed rotation-invariant autoencoder for spherical images. The encoder consists of an $S^2$ convolutional layer followed by a series of $SO(3)$ convolutional layers, and finally the pooling layer which creates a rotation-invariant representation. We can either use this as the encoder output (as in the figure) or use further layers in the encoder to get the output $\mathbf{z}$. The decoder first maps $\mathbf{z}$ to a function on $SO(3)$, using a fully connected layer. This is followed by a stack of $SO(3)$ convolutional layers where all the feature maps are still functions on $SO(3)$. Finally, we use an integration layer to integrate out one of the angles to form a function on $S^2$ which is also the output of the rotation-invariant autoencoder. This framework is trained using the rotation-invariant loss function proposed in this paper.}
    \label{fig:block_diagram}
    \vspace{-0.1in}
\end{figure}

In this section, we describe the proposed autoencoder architecture for signals on spheres, the rotation-invariant latent space, and the novel loss function used to train the network. See Figure \ref{fig:block_diagram} for an illustration of the proposed framework.

\subsection{Encoder}
Given an input spherical image $f^{(0)}: S^2 \to \mathbb{R}^{C_0}$, where $C_0$ is the number of channels in the input, it is passed through multiple stacked convolutional layers each followed by non-linearities. This is similar to the usual convolutional encoder for regular images. The first convolutional layer performs a correlation on $S^2$, between $f^{(0)}$ and the filters $\{h_i^{(1), i=1,\dots,C_1} \}$, where $h_i^{(1)}: S^2 \to \mathbb{R}^{C_0}$. We perform spherical correlation between the input spherical image and the first layer filters, and apply a non-linearity producing the output of the first layer -- a signal $f^{(1)}: SO(3) \to \mathbb{R}^{C_1}$, as explained in Section \ref{sec:background}. The rest of the convolutional layers in the encoder are $SO(3)$ correlations where the filters of layer $d, d>1$,  $\{h_i^{(d)}, i=1,\dots,C_d \}$, where $h_i^{(d)}: SO(3) \to \mathbb{R}^{C_{d-1}}$. After the input is fed through the $D$ convolutional layers producing the feature map $f^{(D)}: SO(3) \to \mathbb{R}^{C_D}$. We now use an integration layer to create the rotation-invariant $f_{inv} \in \mathbb{R}^D$ given by

\begin{equation}
\label{eq:rot_inv}
    f_{inv}(i) = \int_{SO(3)} f_D(X)dX, \quad i=1,\dots,D.
\end{equation}

Once we have a rotation-invariant representation, we use further fully connected layers to map to the latent rotation-invariant representation which we denote by $\mathbf{z} \in \mathbb{R}^N$.

\subsection{Decoder}
The purpose of the decoder is to map the latent representation back to the original input space. To this end, the first layer of the decoder is a fully connected layer that maps $\mathbf{z}$ to a function $f$ on $SO(3)$. We then use a series of $SO(3)$ convolutional layers, similar to those in the encoder, to map to $g: SO(3) \to \mathbb{R}^{C_0}$. Note that $SO(3)$ correlations produce outputs which are functions on $SO(3)$. In order to construct the desired function on $S^2$, we use another integration layer which integrates over $\gamma$:

\begin{equation}
    \hat{f} (\alpha, \beta) = \int_{\gamma = 0}^{2\pi} g(\alpha, \beta, \gamma) d\gamma
\end{equation}

The filters in the various $S^2$ and $SO(3)$ convolutional layers are the trainable parameters in the autoencoders, for which we have to design a novel loss function.

\subsection{Rotation-invariant loss function}
The latent space of the autoencoder is, by design, invariant to 3D rotations of the input spherical images. However, this also means that all the information about the rotation is discarded, and hence the \textit{reconstruction error can only be measured up to an unknown 3D rotation}. To this end, we use the following function

\begin{equation}
    L(f, \hat{f}) = \int_{S^2} |f(x)|^2dx + \int_{S^2} |\hat{f}(x)|^2dx - 2\max_{SO(3)} \{ (f \star \hat{f}) \}  \},
\end{equation}

where the max operator is over the $SO(3)$ function that is the output of the spherical cross-correlation function $\star$. Essentially, the function finds the best rotation $R \in SO(3)$ such that the difference between $f(R^{-1}x)$ and $\hat{f}(x)$. Clearly, the loss function is also rotation-invariant, as it solves for the optimal alignment before computing the usual squared error loss. While training the network, we compute this loss over a mini-batch and average it, as is normally done. We can employ the Fourier correlation theorem to compute the above spherical cross-correlation efficiently. 

\section{Experimental results}
We conduct extensive experiments on the proposed rotation-invariant autoencoder using publicly available datasets. We show that compared to a vanilla autoencoder trained without using the proposed loss function, the proposed framework is significantly better in terms of reconstruction performance, as well as clustering and classification metrics.

\subsection{Spherical MNIST}

\paragraph{Dataset details:} MNIST is a commonly used dataset~\cite{lecun1998mnist} in computer vision research consisting of handwritten images of digits 0-9. Here, we use a spherical version of MNIST, called spherical MNIST, by Cohen et al.~\cite{cohen2018spherical}. Using stereographic projection, all the digits are projected onto the northern hemisphere of a unit sphere. There are two versions of the dataset -- (a) NR: all the digits on the sphere are aligned, i.e. they have the same orientation and (b) R: after projection onto the sphere, the spherical images are rotated randomly. Each of these datasets consists of 60000 training and 10000 test examples. Note that the bandwidth of the signals is set to $b=30$. This means that the size of the spherical signal is $60 \times 60$, i.e., the spherical signal is sampled at 60 values of $\alpha \in [0, 2\pi]$ and $\beta \in [0, \pi]$ using the Driscoll-Healy grid~\cite{driscoll1994computing}.

\paragraph{Autoencoder architecture and training details:} The first layer of the encoder consists of an $S^2$ convolutional layer with $20$ output channels and it reduces the bandwidth to $b=12$. This is followed by a $SO(3)$ convolutional layer with $40$ output channels which further reduces the bandwidth of the feature maps to $6$. Both these layers utilize element-wise ReLU non-linearities as the activation function. As shown in Figure \ref{fig:block_diagram}, we then use the pooling layer defined in Equation \eqref{eq:rot_inv} to create the rotation invariant representation. For most of the experiments, we set the latent dimension, $dim(\mathbf{z}) = 120$. In the decoder, we use a fully connected layer to map $\mathbf{z}$ to a function on $SO(3)$ with $b=6$. This is followed by $2$ $SO(3)$ convolutional layers which upsample the bandwidth to $12$ and $30$. The integration layer is used to produce the reconstructed image on the sphere. We train two types of networks -- (a) Vanilla AE (the baseline) Here, the spherical autoencoder trained using the regular L2 loss given by $L_{euc}(f, \hat{f}) = ||f - \hat{f}||_2^2$ and (b) Rotation-invariant AE trained using the proposed rotation-invariant loss function. These networks are trained for three different settings -- (a) Train NR/Test NR: both the training and test sets are aligned and have the same orientation (b) Train R/Test R: Both the train and test set have random rotations applied to them and (c) Train NR/Test R: The training set is aligned while the test set has random rotations applied to it. That is, in this setting, the trained network is subject to unseen rotations at test time, and is the most challenging setting. The autoencoders are trained for $20$ epochs using Adam optimizer~\cite{kingma2014adam} with an initial learning rate of $0.1$ and a batch size of $32$. 

\begin{figure*}
     \centering
     \begin{subfigure}[b]{\textwidth}
         \centering
         \includegraphics[width=\textwidth]{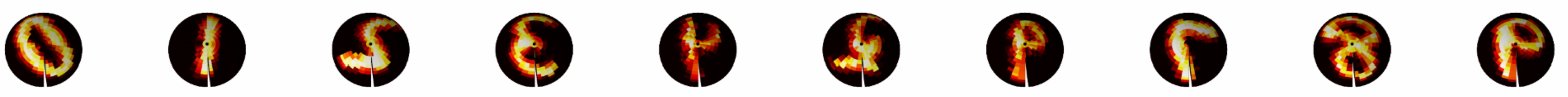}
         \caption{Ground-truth spherical images}
     \end{subfigure}
     \hfill
     \begin{subfigure}[b]{\textwidth}
         \centering
         \includegraphics[width=\textwidth]{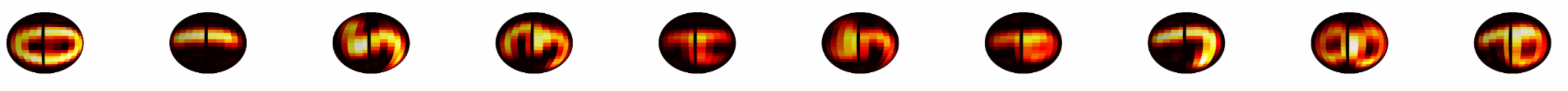}
         \caption{Reconstructed spherical images}
     \end{subfigure}
     \caption{Visualization of the ground-truth spherical images (a) which are also the input to the autoencoder and the reconstructed images (b) from the proposed rotation-invariant autoencoder trained using the rotation-invariant loss function. Note that due to rotation-invariance in the latent space, the reconstruction can be achieved only up to a 3D rotation, as seen in the figure. The spherical images have been partially aligned for easier visualization.}
     \label{fig:mnist_reconstruction}
\end{figure*}

\begin{figure*}
    \centering
    \includegraphics[width=\textwidth,trim={10cm 1cm 9cm 1cm},clip]{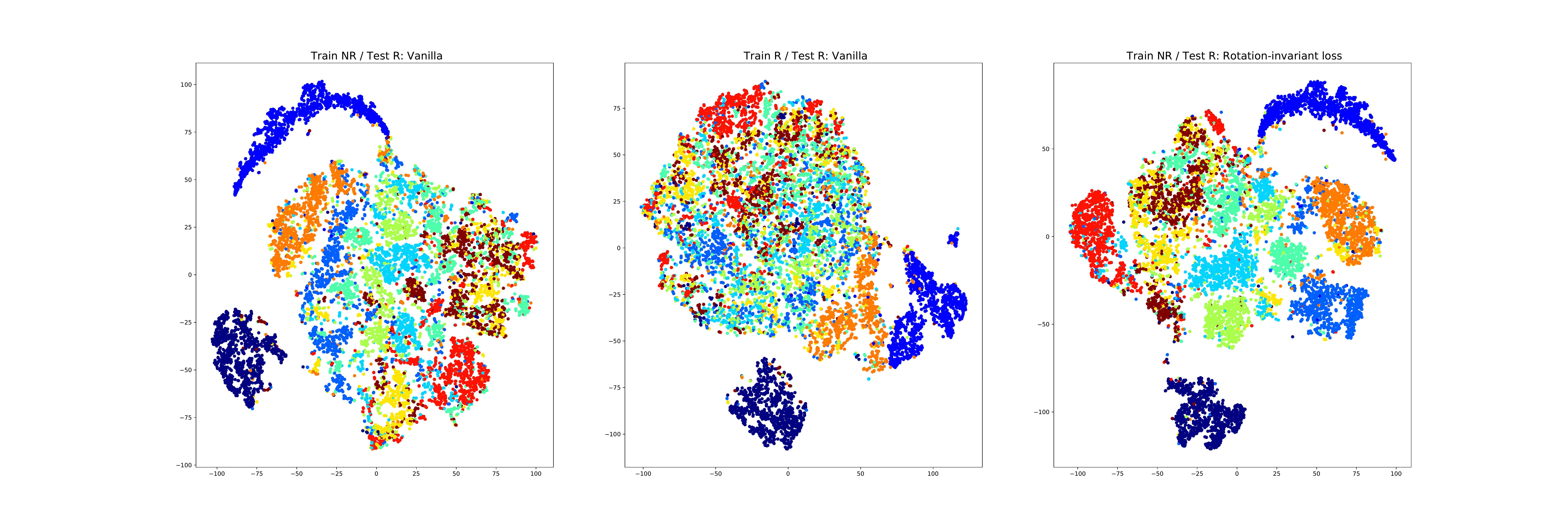}
    \caption{t-SNE visualization of latent features obtained from different spherical autoencoder variants we trained on Spherical MNIST for our experiments. Legend: dark blue = 0, blue = 1, light blue = 2, cyan =3, dark green = 4, light green = 5, yellow = 6, orange = 7, red = 8, brown = 9. As expected, we see a signficant overlap between the points for '6' and '9'.}
    \label{fig:mnist_tsne}
    \vspace{0.1in}
\end{figure*}

\begin{table}
\small
\centering
\vspace{0.5em}
    \begin{tabular}{lcccc}
    \toprule
    Train/test setting & \makecell{Vanilla AE \\(L2 loss)} & \makecell{Proposed AE \\ Rotation-invariant loss} \\
    \midrule
    NR/NR & 18.24 & 22.46 \\
    R/R   & 10.35 & 22.52\\
    NR/R  & 14.33 & 22.41\\
    \bottomrule
\end{tabular}
\caption{Reconstruction results on Spherical MNIST, in terms of PSNR (dB), for different settings and variants of spherical autoencoders. We see that the proposed method outperforms the vanilla AE for all settings. We also observe that the proposed framework yield stable performance over all the settings as expected, as the latent representations are exactly rotation-invariant.}
\label{tab:data_augment_schemes}
\end{table}

\begin{table*}
\small
\centering
\vspace{0.5em}
    \begin{tabular}{lcccc}
    \toprule
    Method & Purity & Homogeneity & Completeness & Classification Accuracy (\%) \\
    \midrule
    Vanilla AE Train NR/Test R                      & 0.37 & 0.27 & 0.29 & 86.58\\
    Vanilla AE Train R/Test R                        & 0.35 & 0.25 & 0.27 & 74.37\\
    Rotation-invariant AE (proposed) Train NR/Test R & \textbf{0.40} & \textbf{0.39} & \textbf{0.31} & \textbf{89.42}\\
    \bottomrule
\end{tabular}
\caption{Clustering and classification results on Spherical MNIST. Clearly, the proposed method outperforms the other unsupervised baselines in all metrics. By comparison, our implementation of $S^2$CNN (described in the text) ~\cite{cohen2018spherical}, which is a fully supervised deep classifier, yields $94\%$ classification accuracy.}
\label{tab:mnist_results}
\end{table*}

\begin{table}
\small
\centering
\setlength\tabcolsep{4pt} 
\vspace{0.5em}
    \begin{tabular}{lcccc}
    \toprule
    $dim(\mathbf{z})$ & \makecell{Vanilla AE \\ Train NR/Test NR} & \makecell{Vanilla AE \\ Train R/Test R} & \makecell{Rot-inv AE \\ Train NR/Test R} \\
    \midrule
    60  & 17.45 & 10.36 & 22.46 \\
    120 & 18.24 & 10.33 & 22.55 \\
    240 & 18.03 & 10.38 & 22.41 \\
    \bottomrule
\end{tabular}
\caption{Effect of dimension of the latent space measured in terms of PSNR (dB) on the test set. We see that the performance is quite stable with respect to $dim(\mathbf{z})$ and using $120$ dimensions yields the best results.}
\label{tab:mnist_latent_dim}
\end{table}

\begin{table}
\small
\centering
\setlength\tabcolsep{4pt} 
\vspace{0.5em}
    \begin{tabular}{ccc}
    \toprule
    \makecell{Percentage of \\ examples labeled}  & \makecell{Deep $S^2$CNN \\ (fully supervised)} & \makecell{Rot-inv AE features \\ (unsupervised)} \\
    \midrule
    1   & 33.90\% & \textbf{63.73\%} \\
    2   & 62.74\% & \textbf{69.28\%} \\
    5   & \textbf{80.82\%} & 74.76\% \\
    10  & \textbf{86.32\%} & 79.71\% \\
    100 & \textbf{95.74\%} & 89.42\% \\
    \bottomrule
\end{tabular}
\caption{Classification accuracy on the test set with few-shot learning. When there are few labeled examples, the unsupervised features can outperform the fully supervised deep $S^2$CNN.}
\label{tab:mnist_few_shot}
\end{table}

\paragraph{Results:} We conduct three sets of experiments on this dataset for the different settings described above.  First, we compare the various methods based on the reconstruction peak signal-to-noise ratio (PSNR) between the input spherical image and the reconstructed image for all the examples in the test set. The results are shown in Table \ref{tab:mnist_results}. We clearly see that the proposed autoencoder trained using the rotation-invariant loss function outperforms the vanilla autoencoder, trained with just the L2 loss, for all cases. It is especially important to note that, compared to Train NR/Test NR for the case of Train NR/Test R, the vanilla autoencoder (trained with the L2 loss) completely fails, while rotation-invariant autoencoder's performance is not affected. 

Second, we gather the latent features for both the train and test images for the various settings using both the vanilla and the proposed rotation-invariant autoencoder. Then, we train a simple multi-class linear classifier using a single fully-connected layer to map from the $120$-dimensional latent feature to a probability distribution over the $10$ classes. We measure the classification accuracy on the held out test set. As shown in Table \ref{tab:mnist_results}, we can see that the proposed framework is significantly better than the vanilla spherical autoencoder. We also compare our method with the purely supervised approach of Cohen et al.~\cite{cohen2018spherical}. For this, we construct a deeper spherical CNN classifier with a $S^2$ convolutional layer followed by a $SO(3)$ convolutional layer, the pooling layer and a fully connected layer. We observe that the features obtained by unsupervised learning actually have a significant amount of discriminative information and perform only a little worse than the fully supervised technique. We also conduct a few-shot learning experiment to better illustrate the advantage of unsupervised feature learning. We see from Table \ref{tab:mnist_few_shot} that when we have very few labeled examples for training classifiers, using the latent features of the proposed rotation-invariant autoencoder outperforms the supervised deep spherical CNN.  

Third, we use the latent features for the various cases and perform both 2D visualization of the latent features using t-SNE~\cite{maaten2008visualizing} as well as $k$-means clustering with $k=10$. The t-SNE plots are shown in Figure \ref{fig:mnist_tsne}, which show that the clusters are clearly more compact and homogeneous for the proposed method, compared to the vanilla autoencoder. Using the clusters from the $k$-means algorithm, we compute common clustering metrics -- purity, homogeneity and completeness for the latent features and report them in Table \ref{tab:mnist_results}, where we observe improved performance using the proposed method. 

\paragraph{Effect of latent dimension} In order to study the effect of the dimension of the latent space, we train autoencoders for the various settings with $dim(\mathbf{z}) = 60, 120, 240$ and record the reconstruction PSNRs on the test set. The results thus obtained are shown in Table \ref{tab:mnist_latent_dim}.

\begin{figure}[]
    \centering
    \includegraphics[width=\linewidth,trim={0 4cm 0 4cm},clip]{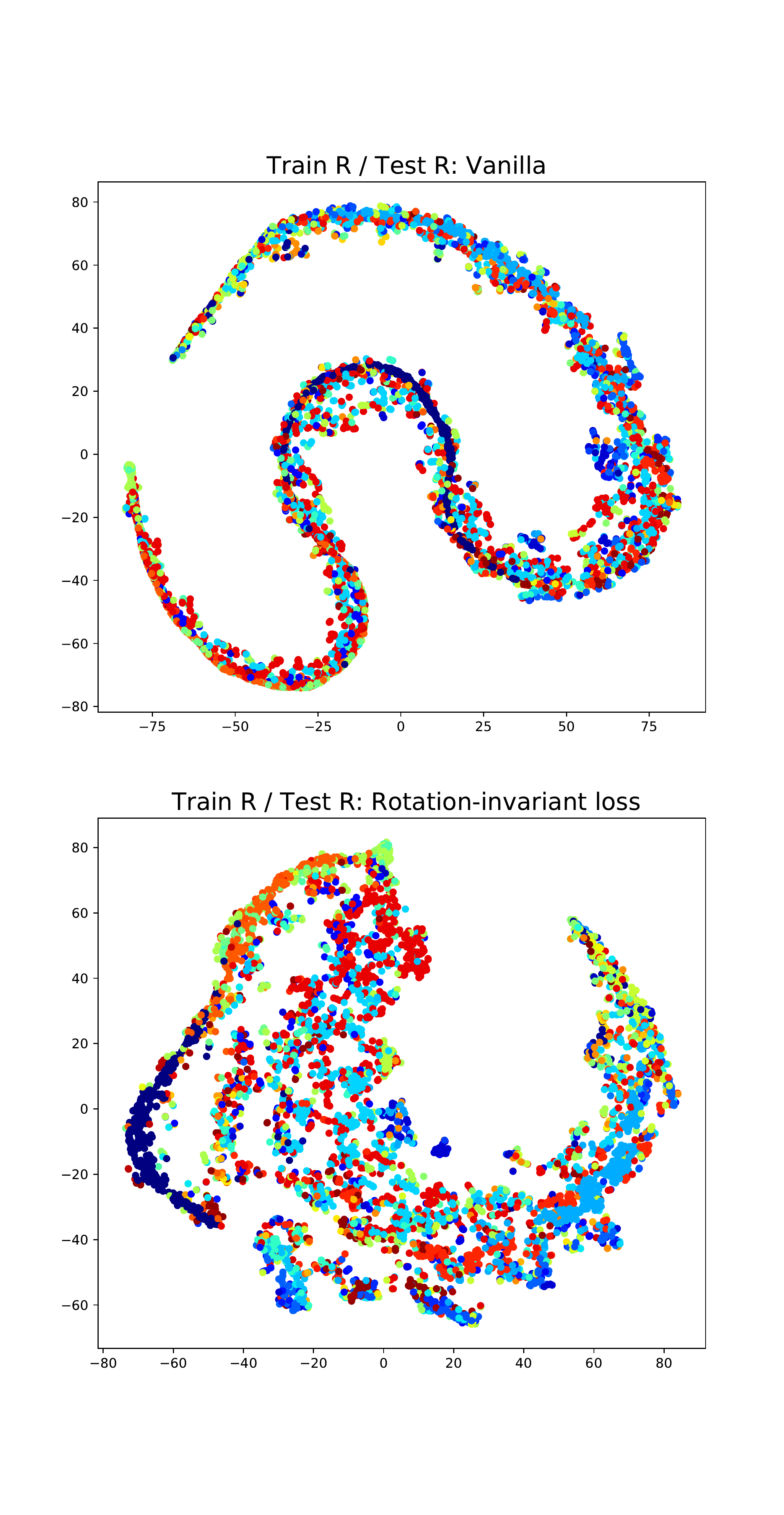}
    \caption{t-SNE visualization of latent features for the SHREC17 dataset. The clusters are more easily visible using the proposed method.}
    \label{fig:shrec_tsne}
\end{figure}

\begin{table*}[!htb]
\small
\centering
\vspace{0.5em}
    \begin{tabular}{lcccc}
    \toprule
    Method & Purity & Homogeneity & Completeness & Classification Accuracy (\%) \\
    \midrule
    Vanilla AE  & 0.30 & 0.26 & 0.22 & 48.80\\
    Rotation-invariant AE (proposed) & 0.41 & 0.38 & 0.31 & 57.76\\
    \bottomrule
\end{tabular}
\caption{Clustering and classification results on SHREC17. Clearly, the proposed method outperforms the other unsupervised baselines in all metrics. By comparison, our implementation of $S^2$CNN (with 3 conv layer) ~\cite{cohen2018spherical}, which is a fully supervised deep classifier, yields $65.42\%$ classification accuracy.}
\label{tab:shrec_clustering}
\end{table*}

\begin{table*}[!htb]
    \centering
    \begin{tabular}{llllllll}
    \toprule
   Type & Method           & P@N & R@N   & F1@N  & mAP   & NDCG  \\ \hline
   \multirow{5}{*}{Supervised} & Tatsuma\_ReVGG   & 0.705  & 0.769 & 0.719 & 0.696 & 0.783 \\ 
   & Furuya\_DLAN     & 0.814 & 0.683 & 0.706 & 0.656 & 0.754 \\ 
   & SHREC16-Bai\_GIFT &0.678& 0.667 & 0.661 & 0.607 & 0.735 \\ 
   & Deng\_CM-VGG5-6DB &0.412& 0.706 & 0.472 & 0.524 & 0.624 \\
   & $S^2$CNN \cite{cohen2018spherical}   & 0.701 & 0.711 & 0.699 & 0.676 & 0.756\\
   \midrule
    \multirow{2}{*}{Unsupervised} & Vanilla AE  & 0.075 & 0.092 & 0.066 & 0.008 & 0.064 \\
    & Rotation-invariant AE (proposed) & 0.351 & 0.361 & 0.335 & 0.215 & 0.345\\
    \bottomrule
    \end{tabular}
    \caption{\label{tab:shrec_retrieval} Shape retrieval results on SHREC17 compared to fully supervised methods. We see that even the unsupervised features obtained using our proposed method with the rotation-invariant loss function yields good results, especially when compared to the vanilla spherical autoencoder with the L2 loss function. The results for the supervised methods are taken from \cite{cohen2018spherical}.}
\end{table*}

\subsection{SHREC17 shape retrieval}

\paragraph{Dataset details:} The dataset consists~\cite{savva2016shrec16} of $51300$ 3D meshes belonging to $55$ classes such as chairs, tables and airplanes. For the experiments, we use the variant of the dataset where both the training and test set are randomly rotated. The 3D meshes are centered at the origin and are projected onto a sphere around the mesh using ray casting. Rays are cast from the origin passing through the mesh and hitting the sphere. The distance between the point on the mesh and the sphere is recorded as the signal on the sphere, and forms the spherical representation of the 3D mesh. As before Driscoll-Healy grid is used to sample the sphere with a bandwidth $b=30$, i.e. the spherical signals can be stored as $60 \times 60$ arrays.

\paragraph{Autoencoder architecture and training details:} Both the autoencoder architecture and the hyperparameters are the same as in the case of the Spherical MNIST dataset. As before, the latent space is $120$-dimensional and invariant to 3D rotations applied at the input. As before, we train a vanilla spherical autoencoder trained using a simple L2 loss function and the proposed autoencoder framework trained using the rotation-invariant loss function. 

\paragraph{Results:} We conduct three set of experiments on this dataset. First, we compute the latent features for both the training and test set 3D shape spherical representations. The latent features are then used to train a simple linear classifier. The results are shown in Table \ref{tab:shrec_clustering}. We see that the proposed framework yields significantly higher classification accuracy, compared to vanilla AE. Second, as in the case of Spherical MNIST, we conduct similar clustering experiments by performing $k$-means clustering with $k=55$. We report commonly used clustering metrics -- purity, homogeneity and completeness -- in Table \ref{tab:shrec_clustering}. We can clearly observe improved performance with the proposed method, compared to the vanilla AE trained with L2 loss. Third, we conduct shape retrieval on the held out test set following the protocol in \cite{savva2016shrec16}. The results are reported in Table \ref{tab:shrec_retrieval} for various retrieval metrics and comparison with fully supervised methods is also provided. We observe that the features obtained through the proposed method work reasonably well for shape retrieval. 

\section{Conclusion}
In this paper, we presented a novel architecture for learning rotation-invariant representations for spherical images in an unsupervised fashion. We designed an autoencoder using $S^2$ and $SO(3)$ convolutional layers, which are equivariant to $3D$ rotations and the latent space is constrained to be rotation-invariant using a pooling layer. We proposed a rotation-invariant loss function based on the maximum of spherical cross-correlation in order to train such networks. Through experiments on multiple datasets of spherical images and spherical representations of 3D shapes, we showed the utility of such representations. The proposed method results in good performance in terms of clustering, retrieval and classification based on unsupervised learning. We hope that this work will lead to similar transformation-invariant unsupervised feature learning methods for other transformations such as scale, affine transforms etc. and for other domains such as images, graphs and sets.

{\small
\bibliographystyle{ieee_fullname}
\bibliography{egbib}
}

\end{document}